\documentclass[Afour,times,sageh]{sagej}
\usepackage{times}

\usepackage{natbib}
\usepackage{hyperref}

\usepackage[nolist,nohyperlinks]{acronym}
\usepackage{graphicx}
\usepackage{epstopdf}
\usepackage{epsfig}
\usepackage{algorithm}
\usepackage{algpseudocode}
\usepackage[disable]{todonotes} 
\usepackage{multirow}
\usepackage{flushend}
\usepackage{amsmath}
\usepackage[font=small,labelfont=bf]{caption}
\usepackage[squaren]{SIunits}
\usepackage{comment}
\usepackage{subcaption}

\newcommand{\segmap}{\textit{SegMap}}

\def\eg{\emph{e.g.\ }}
\def\ie{\emph{i.e.\ }}

\begin{document}
\par\smallskip\noindent
\begin{minipage}{\textwidth}
\begin{center}
This paper has been accepted for publication in the International Journal of Robotics Research.
\vspace{12pt}

DOI: 10.1177/0278364919863090
\vspace{30pt}

Please cite our work as:
\vspace{12pt}

R. Dub\'e, A. Cramariuc, D. Dugas, H. Sommer, M. Dymczyk, J. Nieto, R. Siegwart, C. Cadena. ``SegMap: Segment-based mapping and localization using data-driven descriptors''. \textit{The International Journal of Robotics Research.}
\vspace{30pt}
\end{center}

bibtex:
\begin{verbatim}
@article{segmap2019dube,
  author    = {Renaud Dub\'e and Andrei Cramariuc and Daniel Dugas
               and Hannes Sommer and Marcin Dymczyk and Juan Nieto
               and Roland Siegwart and Cesar Cadena},
  title     = {{SegMap}: Segment-based mapping and localization using
               data-driven descriptors},
  journal   = {The International Journal of Robotics Research},
  doi       = {10.1177/0278364919863090}
}
\end{verbatim}
\end{minipage}
\par\smallskip

\newpage

\title{SegMap: Segment-based mapping and localization using data-driven descriptors}

\author{Renaud Dub\'e\affilnum{1}\affilnum{2}\affilnum{*}, Andrei Cramariuc\affilnum{1}\affilnum{*}, Daniel Dugas\affilnum{1}, Hannes Sommer\affilnum{1}\affilnum{2}, Marcin Dymczyk\affilnum{1}\affilnum{2}, Juan Nieto\affilnum{1}, Roland Siegwart\affilnum{1}, and Cesar Cadena\affilnum{1}}
\affiliation{\affilnum{1}Autonomous Systems Lab (ASL), ETH Zurich, Switzerland\\
\affilnum{2}Sevensense Robotics AG, Zurich, Switzerland\\
\affilnum{*}The authors contributed equally to this work.}

\corrauth{Renaud Dub\'e and Andrei Cramariuc}
\email{renaud.dube@sevensense.ch and crandrei@ethz.ch}

\begin{abstract}

Precisely estimating a robot's pose in a prior, global map is a fundamental capability for mobile robotics, \eg autonomous driving or exploration in disaster zones.
This task, however, remains challenging in unstructured, dynamic environments, where local features are not discriminative enough and global scene descriptors only provide coarse information.
We therefore present \segmap{}: a \textit{map representation} solution for localization and mapping based on the extraction of segments in 3D point clouds.
Working at the level of segments offers increased invariance to view-point and local structural changes, and facilitates real-time processing of large-scale 3D data.
\segmap{} exploits a single compact data-driven descriptor for performing multiple tasks: global localization, 3D dense map reconstruction, and semantic information extraction.
The performance of \segmap{} is evaluated in multiple urban driving and search and rescue experiments.
We show that the learned \segmap{} descriptor has superior segment retrieval capabilities, compared to state-of-the-art handcrafted descriptors.
In consequence, we achieve a higher localization accuracy and a 6\% increase in recall over state-of-the-art.
These segment-based localizations allow us to reduce the open-loop odometry drift by up to 50\%.
\segmap{} is open-source available along with easy to run demonstrations.
\end{abstract}


\maketitle

\section{Introduction}

Mapping and localization are fundamental competencies for mobile robotics and have been well-studied topics over the last couple of decades~(\cite{Cadena16tro-SLAMfuture}).
Being able to map an environment and later localize within it unlocks a multitude of applications, that include autonomous driving, rescue robotics, service robotics, warehouse automation or automated goods delivery, to name a few.
Robotic technologies undoubtedly have the potential to disrupt those applications within the next years.
In order to allow for the successful deployment of autonomous robotic systems in such real-world environments, several challenges need to be overcome: mapping, localization and navigation in difficult conditions, for example crowded urban spaces, tight indoor areas or harsh natural environments.
Reliable, prior-free global localization lies at the core of this challenge.
Knowing the precise pose of a robot is necessary to guarantee reliable, robust and most importantly safe operation of mobile platforms and also allows for multi-agent collaborations.

The problem of mapping and global localization has been well covered by the research community.
On the one hand, a large body of algorithms use cameras and visual cues to perform place recognition.
Relying purely on appearance has, however, significant limitations.
In spite of tremendous progress within this field, state-of-the-art algorithms still struggle with changing seasons, weather or even day-night variations~(\cite{lowryvisual}).
On the other hand, several approaches address the variability of appearance by relying instead on the 3D structure extracted from LiDAR data, which is expected to be more consistent across the aforementioned changes.
Current LiDAR-based \ac{SLAM} systems, however, mostly use the 3D structure for local odometry estimation and map tracking, but fail to perform global localization without any prior on the pose of the robot~(\cite{Hess16}).
%

There exist several approaches that propose to use 3D point clouds for global place recognition.
Some of them make use of various local features~(\cite{rusu2009fast, salti2014shot}), which permit to establish correspondences between a query scan and a map and subsequently estimate a 6-\ac{DoF} pose.
The performance of those systems is limited, as local features are often not discriminative enough and not repeatable given the changes in the environment.
Consequently, matching them is not always reliable and also incurs a large computational cost given the number of processed features.
Another group of approaches relies on global descriptors of 3D LiDAR scans~(\cite{yin2018locnet}) that permit to find a correspondence in the map.
Global descriptors, however, are view-point dependent, especially when designed for only rotational-invariance and not as translation-invariant.
Furthermore, a global scan descriptor is more prone to failures under dynamic scenes, \eg parked cars, which can be important for reliable global localization in crowded, urban scenarios.

\begin{figure}
\centering
\includegraphics[width=1.0\columnwidth]{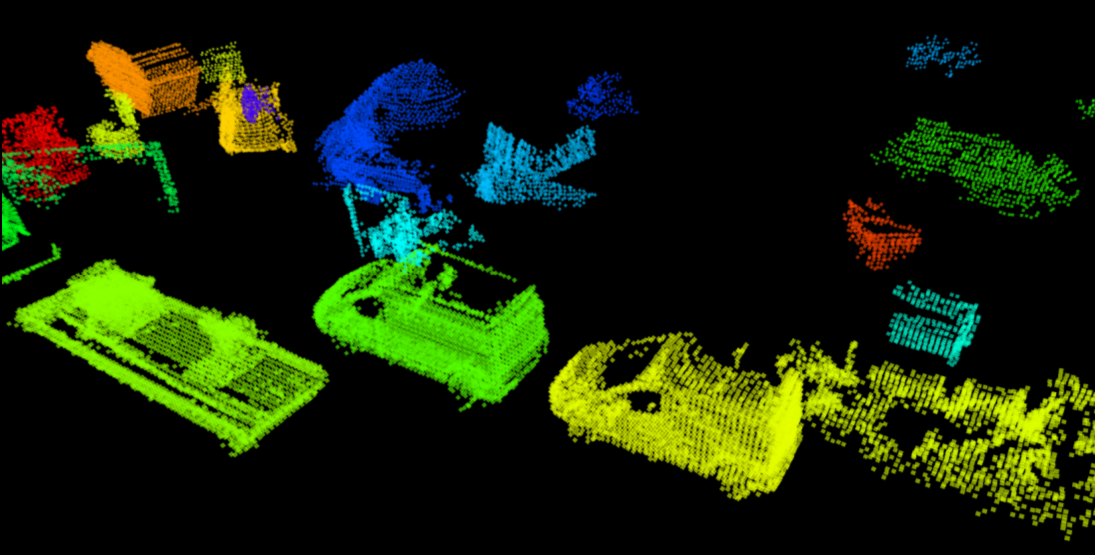}
\caption{Exemplary segments extracted from 3D LiDAR data collected in a rural environment.
These segments were extracted with an incremental Euclidean distance-based region-growing algorithm and represent, among others, vehicles, vegetation and parts of buildings~(\cite{dube2018incremental}).}
\label{fig_segments}
\end{figure}

We therefore present \segmap{}\footnote{\segmap{} is open-source available along with easy to run demonstrations at \url{www.github.com/ethz-asl/segmap}. A video demonstration is available at \url{https://youtu.be/CMk4w4eRobg}}: a unified approach for \textit{map representation} in the localization and mapping problem for 3D LiDAR point clouds.
\segmap{} is formed on the basis of partitioning point clouds into sets of descriptive segments (\cite{dube2017segmatch}), as illustrated in Figure~\ref{fig_segments}.
The segment-based localization combines the advantages of global scan descriptors and local features -- it offers reliable matching of segments and delivers accurate 6-\ac{DoF} global localizations in real-time.
The 3D segments are obtained using efficient region-growing techniques which are able to repeatedly form similar partitions of the point clouds (\cite{dube2018incremental}).
This partitioning provides the means for compact, yet discriminative features to efficiently represent the environment.
During localization global data associations are identified by segment descriptor retrieval, leveraging the repeatable and descriptive nature of segment-based features. 
%
%
%
%
This helps satisfy strict computational, memory and bandwidth constraints, and therefore makes the approach appropriate for real-time use in both multi-robot and long-term applications.

\begin{figure}
\centering
\includegraphics[width=1.0\columnwidth]{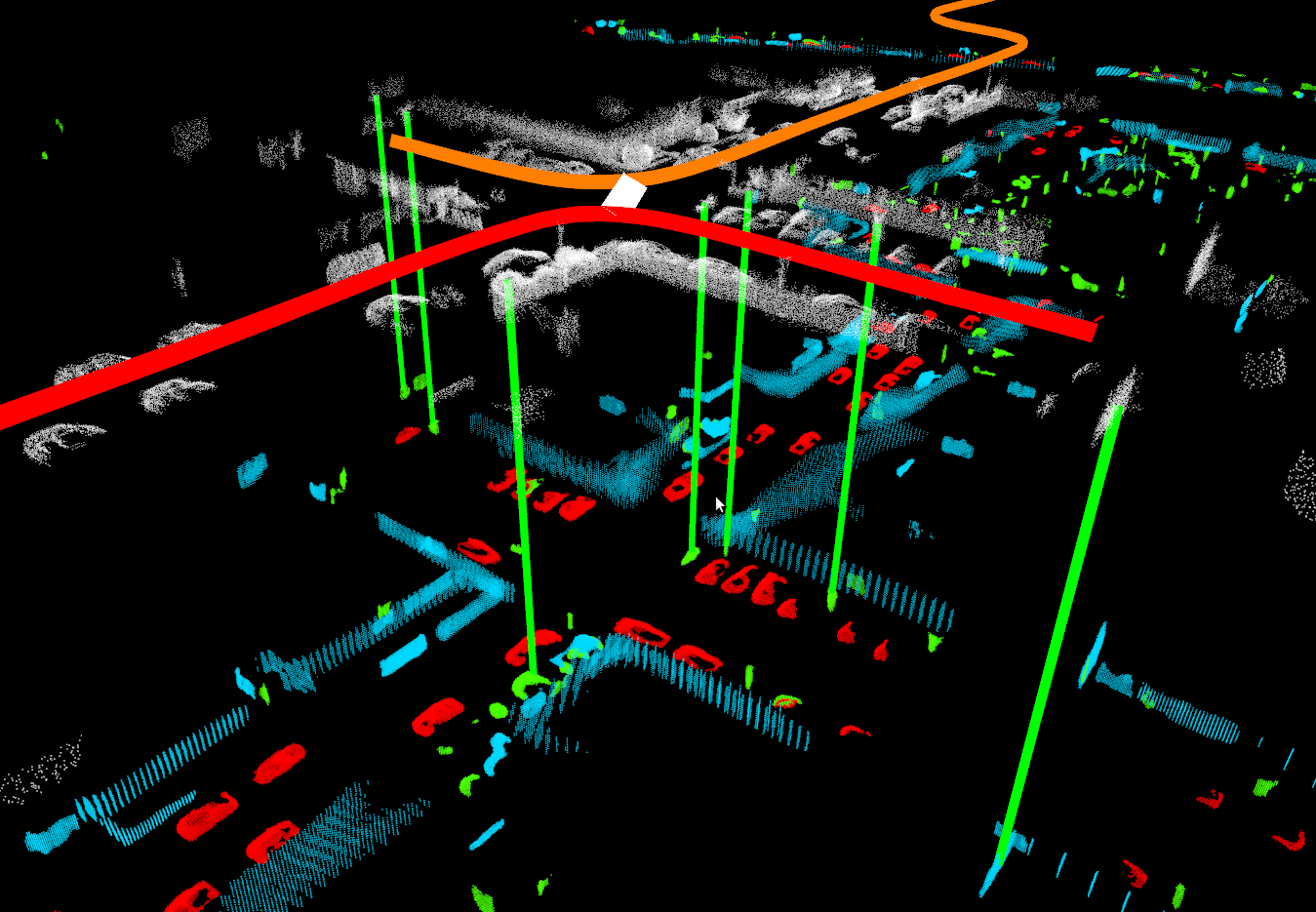}
\caption{An illustration of the \segmap{} approach.
The red and orange paths represent the trajectories of two robots driving simultaneously in opposite directions through an intersection.
In white we show the local segments extracted from the robots' vicinity and characterized using our compact data-driven descriptor.
Correspondences are then made with the target segments, resulting in a successful localization depicted with green vertical lines.
A reconstruction of the target segments is illustrated below, where colors represent semantic information (cars in red, buildings in light blue, and others in green), all possible by leveraging the same compact representation.
We take advantage of the semantic information by performing localization only against static objects, improving robustness against dynamic changes.
Both the reconstruction and semantic classification are computed by leveraging the same descriptors used for global prior-free localization.}
\label{fig_teaser}
\end{figure}

Previous work on segment-based localization considered hand-crafted features and provided only a sparse representation (\cite{dube2017segmatch, tinchev2018}).
These features lack the ability to generalize to different environments and offer very limited insights into the underlying 3D structure.
In this work, we overcome these shortcomings by introducing a novel data-driven segment descriptor which offers high retrieval performance, even under variations in view-point, and that generalizes well to unseen environments.
Moreover, as segments typically represent meaningful and distinct elements that make up the environment, a scene can be effectively summarized by only a handful of descriptors.
The resulting reconstructions, as depicted in Figure~\ref{fig_teaser}, can be built at no extra cost in descriptor computation or bandwidth usage.
They can be used by robots for navigating around obstacles and visualized to improve situational awareness of remote operators.
Moreover, we show that semantic labeling can be executed through classification in the descriptor space.
This information can, for example, lead to increased robustness to changes in the environment by rejecting inherently dynamic classes.

To the best of our knowledge, this is the first work on robot localization that is able to leverage the extracted features for reconstructing environments in three dimensions and for retrieving semantic information.
This reconstruction is, in our opinion, a very interesting capability for real-world, large-scale applications with limited memory and communication bandwidth.
To summarize, this paper presents the following contributions:
\begin{itemize}
\item A data-driven 3D segment descriptor that improves localization performance.
\item A novel technique for reconstructing the environment based on the same compact features used for localization.
\item An extensive evaluation of the \segmap{} approach using real-world, multi-robot automotive and disaster scenario datasets.
\end{itemize}

In relation to the \emph{Robotics: Science and System} conference paper (\citet{dube2018segmap}), we make the following additional contributions:
\begin{itemize}
    \item A comparison of the accuracy of our localization output with the results of recently published technique based on data-driven global 3D scan descriptors (\cite{yin2018locnet}).
    \item An evaluation of trajectory estimates by combining our place recognition approach with a state-of-the-art 3D LiDAR-based SLAM technique (\cite{zhang2014loam}).
    \item A triplet loss descriptor training technique and its comparison to the previously introduced classification-based approach.
    \item A particularly lightweight variant of our \segmap{} descriptor that can be deployed on platforms with limited computational resources.
\end{itemize}

The remainder of the paper is structured as follows: Section~\ref{sec:related_work} provides an overview of the related work in the fields of localization and learning-based descriptors for 3D point clouds.
The \segmap{} approach and our novel descriptor that enables reconstruction of the environment are detailed in Section~\ref{sec:segmap} and Section~\ref{sec:descriptor}.
The method is evaluated in Section~\ref{sec:experiments}, and finally Sections~\ref{sec:discussion}~and~\ref{sec:conclusion} conclude with a short discussion and ideas on future works.

\section{RELATED WORK}
\label{sec:related_work}
This section first introduces state-of-the-art approaches to localization in 3D point clouds.
Data driven techniques using 3D data which are relevant to the present work are then presented.

\textbf{Localization in 3D point clouds}\hspace{2pt}  Detecting loop-closures from 3D data has been tackled with different approaches.
We have identified three main trends: (i) approaches based on local features, (ii) global descriptors and (iii) based on planes or objects.

A significant number of works propose to extract local features from keypoints and perform matching on the basis of these features.
\citet{bosse2013place} extract keypoints directly from the point clouds and describe them with a 3D \textit{Gestalt} descriptor.
Keypoints then vote for their nearest neighbors in a \textit{vote matrix} which is eventually thresholded for recognizing places.
A similar approach has been used in \cite{Gawel2016}.
Apart from such Gestalt descriptors, a number of alternative local feature descriptors exist, which can be used in similar frameworks.
This includes features such as Fast Point Feature Histogram (FPFH)~(\cite{rusu2009fast}) and SHOT~(\cite{salti2014shot}).
Alternatively, \citet{zhuang20133} transform the local scans into bearing-angle images and extract Speeded Up Robust Features (SURFs) from these images.
A strategy based on 3D spatial information is employed to order the scenes before matching the descriptors.
A similar technique by \citet{steder2010robust} first transforms the local scans into a range image.
Local features are extracted and compared to the ones stored in a database, employing the Euclidean distance for matching keypoints.
This work is extended in \cite{steder2011place} by using Normal-Aligned Radial Features (NARF) descriptors and a bag of words approach for matching.


Using global descriptors of the local point cloud for place recognition is also proposed in (\cite{rohling2015fast,granstrom2011learning,magnusson2009automatic, cop2018delight}).
\citet{rohling2015fast} propose to describe each local point cloud with a 1D histogram of point heights, assuming that the sensor keeps a constant height above the ground.
The histograms are then compared using the \textit{Wasserstein} metric for recognizing places.
\citet{granstrom2011learning} describe point clouds with rotation invariant features such as volume, nominal range, and range histogram.
Distances are computed for feature vectors and cross-correlation for histogram features, and an AdaBoost classifier is trained to match places.
Finally, \ac{ICP} is used for computing the relative pose between point clouds.
In another approach, \citet{magnusson2009automatic} split the cloud into overlapping grids and compute shape properties (spherical, linear, and several type of planar) of each cell and combine them into a matrix of surface shape histograms.
Similar to other works, these descriptors are compared for recognizing places.
Recently, \cite{cop2018delight} proposed to leverage LiDAR intensity information with a global point cloud descriptor.
A two-stage approach is adopted such that, after retrieving places based on global descriptors retrieval, a local keypoint-based geometric verification step estimates localization transformations.
The authors demonstrated that using intensity information can reduce the computational timings.
However, the complete localization pipeline operates at a frequency one order of magnitude lower than most LiDAR sensor frequencies.

While local keypoint features often lack descriptive power, global descriptors can struggle with variations in view-point.
Therefore other works have also proposed to use 3D shapes or objects for the place recognition task.
\citet{fernandez2013fast}, for example, propose to perform place recognition by detecting planes in 3D environments.
The planes are accumulated in a graph and an interpretation tree is used to match sub-graphs.
A final geometric consistency test is conducted over the planes in the matched sub-graphs.
The work is extended in \cite{fernandez2016scene} to use the covariance of the plane parameters instead of the number of points in planes for matching.
This strategy is only applied to small, indoor environments and assumes a plane model which is no longer valid in unstructured environments.
A somewhat analogous, seminal work on object-based loop-closure detection in indoor environments using RGB-D cameras is presented by \citet{finman2015icraws}.
Although presenting interesting ideas, their work can only handle a small number of well segmented objects in small scale environments.
Similarily, \citet{bowman2017probabilistic} proposed a novel SLAM solution in which semantic information and local geometric features are jointly incorporated into a probabilistic framework. 
Such semantic-based approaches have significant potential, for example robustness to stark changes in point of view, but require the presence of human-known objects in the scene.

We therefore aim for an approach which does not rely on assumptions about the environment being composed of simplistic geometric primitives such as planes, or a rich library of objects.
This allows for a more general, scalable solution.

\textbf{Learning with 3D point clouds}\hspace{2pt} In recent years, \acp{CNN} have become the state-of-the-art-method for generating learning-based descriptors, due to their ability to find complex patterns in data (\cite{krizhevsky2012imagenet}). 
For 3D point clouds, methods based on \acp{CNN} achieve impressive performance in applications such as object detection (\cite{engelcke2017vote3deep, maturana2015voxnet, riegler2016octnet, livehicle, wu20153d, wohlhart2015learning, qi2016pointnet, fang20153d}), semantic segmentation (\cite{riegler2016octnet, livehicle, qi2016pointnet, tchapmi2017segcloud, graham20183d, wu2018squeezeseg}), 
and 3D object generation (\cite{wu2016learning}), and LiDAR-based local motion estimation~(\cite{dewan2018iros, velas2018cnn}). 

Recently, a handful of works proposing the use of \acp{CNN} for localization in 3D point clouds have been published. 
First, \citet{zeng20163dmatch} proposes extracting data-driven 3D keypoint descriptors (3DMatch) which are robust to changes in view-point. 
Although impressive retrieval performance is demonstrated using an RGB-D sensor in indoor environments, it is not clear whether this method is applicable in real-time in large-scale outdoor environments.
A different approach based on 3D \acp{CNN} was proposed in \cite{ye2017place} for performing localization in semi-dense maps generated with visual data.
Recently, \citet{yin2017locnet} introduced a semi-handcrafted global descriptor for performing place recognition and rely on an \ac{ICP} step for estimating the 6-\ac{DoF} localization transformations.
This method will be used as a baseline solution in Section~\ref{loam_experiment} when evaluating the precision of our localization transformations.
\citet{elbaz20173d} propose describing local subsets of points using a deep neural network autoencoder. 
The authors state, however, that the implementation has not been optimized for real-time operation and no timings have been provided.
In contrast, our work presents a data-driven segment-based localization method that can operate in real-time and that enables map reconstruction and semantic extraction capabilities. 


To achieve this reconstruction capability, the architecture of our descriptor was inspired by autoencoders in which an encoder network compresses the input to a small dimensional representation, and a  decoder network attempts to decompress the representation back into the original input.
%
%
The compressed representation can be used as a descriptor for performing 3D object classification (\cite{brock2016generative}).
\citet{brock2016generative}~also present successful results using variational autoencoders for reconstructing voxelized 3D data.
%
Different configurations of encoding and decoding networks have also been proposed for achieving localization and for reconstructing and completing 3D shapes and environments (\cite{guizilini2017learning, dai2016shape, varley2016shape, ricao2017compressed,elbaz20173d,schonberger2017semantic}).

While autoencoders present the interesting opportunity of simultaneously accomplishing both compression and feature extraction tasks, optimal performance at both is not guaranteed.
As will be shown in Section~\ref{ssec:retrieval_performance}, these two tasks can have conflicting goals when robustness to changes in point of view is desired.
In this work, we combine the advantages of the encoding-decoding architecture of autoencoders with a technique proposed by~\citet{parkhi2015deep}.
The authors address the face recognition problem by first training a \ac{CNN} to classify people in a training set and afterwards use the second to last layer as a descriptor for new faces. 
Other alternative training techniques include for example the use of contrastive loss (\cite{bromley1994signature}) or triplet loss (\cite{weinberger2006distance}), the latter one being evaluated in Section~\ref{ssec:retrieval_performance}.
We use the resulting segment descriptors in the context of \ac{SLAM} to achieve better performance, as well as significantly compressed maps that can easily be stored, shared, and reconstructed.


\section{The \segmap{} approach}
\label{sec:segmap}
This section presents our \segmap{} approach to localization and mapping in 3D point clouds.
It is composed of five core modules: segment extraction, description, localization, map reconstruction, and semantics extraction.
These modules are detailed in this section and together allow single and multi-robot systems to create a powerful unified representation which can conveniently be transferred. 

%
%

\textbf{Segmentation}\hspace{2pt} The stream of point clouds generated by a 3D sensor is first accumulated in a dynamic voxel grid\footnote{In our experiments, we consider two techniques for estimating the local motion by registering successive LiDAR scans: one which uses \ac{ICP} and one based on LOAM (\cite{zhang2014loam}).}.
Point cloud segments are then extracted in a section of radius $R$ around the robot.
In this work we consider two types of incremental segmentation algorithms~(\cite{dube2018incremental}).
The first one starts by removing points corresponding to the ground plane, which acts as a separator for clustering together the remaining points based on their Euclidean distances. 
The second algorithm computes local normals and curvatures for each point and uses these to extract flat or planar-like surfaces.
Both methods are used to incrementally grow segments by using only newly active voxels as seeds which are either added to existing segments, form new segments or merge existing segments together\footnote{For more information on these segmentation algorithms, the reader is encouraged to consult our prior work~(\cite{dube2018incremental}).}.
This results in a handful of local segments, which are individually associated to a set of past observations \ie ${S_i=\{s_1, s_2, \ldots, s_n \}}$. 
Each observation $s_j \in S_i$ is a 3D point cloud representing a snapshot of the segment as points are added to it.
Note that $s_n$ represents the latest observation of a segment and is considered \textit{complete} when no further measurements are collected, \eg{}when the robot has moved away.

\textbf{Description}\hspace{2pt} Compact features are then extracted from these 3D segment point clouds using the data-driven descriptor presented in Section~\ref{sec:descriptor}.
A global segment map is created online by accumulating the segment centroids and corresponding descriptors.
In order for the global map to most accurately represent the latest state of the world, we only keep the descriptor associated with the last and most complete observation.

\begin{figure*}
\centering
\includegraphics[width=1.0\linewidth]{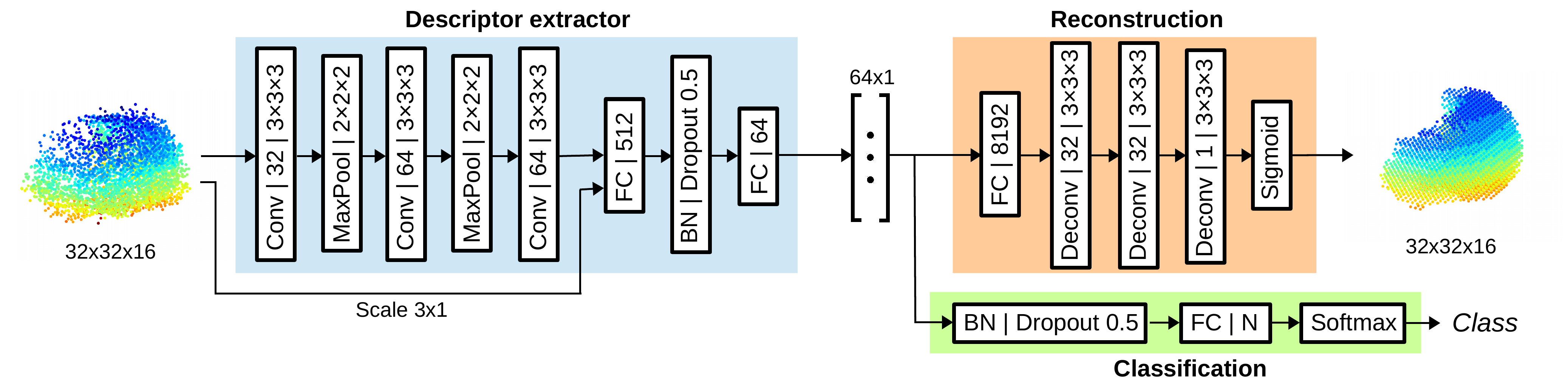}
\caption{
The descriptor extractor is composed of three convolutional and two fully connected layers.
The 3D segments are compressed to a representation of dimension $64\times1$ which can be used for localization, map reconstruction and semantic extraction.
Right of the descriptor we illustrate the classification and reconstruction layers which are used for training.
In the diagram the convolutional (Conv), deconvolutional (Deconv), fully connected (FC) and batch normalization (BN) layers are abbreviated respectively.
As parameters the Conv and Deconv layers have the number of filters and their sizes, FC layers have the number of nodes, max pool layers have the size of the pooling operation, and dropout layers have the ratio of values to drop.
Unless otherwise specified, \ac{ReLU} activation functions are used for all layers.
}
\label{fig:descriptor_architecture}
\end{figure*}

\textbf{Localization}\hspace{2pt} In the next step, candidate correspondences are identified between global and local segments using \ac{k-NN} in feature space. 
The approximate $k$ nearest descriptors are retrieved through an efficient query in a kd-tree.
Localization is finally performed by verifying the largest subset of candidate correspondences for geometrical consistency on the basis of the segment centroids.
Specifically, the centroids of the corresponding local and global segments must have the same geometric configuration up to a small jitter in their position, to compensate for slight variations in segmentation. 
In the experiments presented in Section~\ref{final_experiment}, this is achieved using an incremental recognition strategy which uses caching of correspondences for faster geometric verifications~(\cite{dube2018incremental}).

When a large enough geometrically consistent set of correspondence is identified, a 6-\ac{DoF} transformation between the local and global maps is estimated.
This transformation is fed to an incremental pose-graph \ac{SLAM} solver which in turn estimates, in real-time, the trajectories of all robots (\cite{dube2017multirobot}).

\textbf{Reconstruction}\hspace{2pt} Thanks to our autoencoder-like descriptor extractor architecture, the compressed representation can at any time be used to reconstruct an approximate map as illustrated in Figure~\ref{fig:reconstruction_meshes_kitti}.
As the \segmap{} descriptor can conveniently be transmitted over wireless networks with limited bandwidth, any agent in the network can reconstruct and leverage this 3D information.
More details on these reconstruction capabilities are given in Section~\ref{ssec:training}.

\textbf{Semantics}\hspace{2pt} The \segmap{} descriptor also contains semantically relevant information without the training process having enforced this property on the descriptor.
This can, for example, be used to discern between static and dynamic objects in the environment to improve the robustness of the localization task.
In this work we present an experiment where the network is able to distinguish between three different semantic classes: \textit{vehicles}, \textit{buildings}, and \textit{others} (see Section~\ref{ssec:training_semantics}).

\section{The \segmap{} Descriptor}
\label{sec:descriptor}
In this section we present our main contribution: a data-driven descriptor for 3D segment point clouds which allows for localization, map reconstruction and semantic extraction.
The descriptor extractor's architecture and the processing steps for inputting the point clouds to the network are introduced.
We then describe our technique for training this descriptor to accomplish tasks of both segment retrieval and map reconstruction.
We finally show how the descriptor can further be used to extract semantic information from the point cloud.

\subsection{Descriptor extractor architecture}
The architecture of the descriptor extractor is presented in Figure~\ref{fig:descriptor_architecture}.
Its input is a 3D binary voxel grid of fixed dimension $32\times32\times16$ which was determined empirically to offer a good balance between descriptiveness and the size of the network.
The description part of the \ac{CNN} is composed of three 3D convolutional layers with max pool layers placed in between and two fully connected layers.
Unless otherwise specified, \ac{ReLU} activation functions are used for all layers.
The original scale of the input segment is passed as an additional parameter to the first fully connected layer to increase robustness to voxelization at different aspect ratios.
The descriptor is obtained by taking the activations of the extractor's last fully connected layer.
This architecture was selected by grid search over various configurations and parameters.

\subsection{Segment alignment and scaling}
\label{sec:alignment}
A pre-processing stage is required in order to input the 3D segment point clouds for description.
First, an alignment step is applied such that segments extracted from the same objects are similarly presented to the descriptor network.
This is performed by applying a 2D Principal Components Analysis (PCA) of all points located within a segment.
The segment is then rotated so that the $x$-axis of its frame of reference, from the robot's perspective, aligns with the eigenvector corresponding to the largest eigenvalue.
We choose to solve the ambiguity in direction by rotating the segment so that the lower half section along the $y$-axis of  its frame of reference contains the highest number of points.
From the multiple alignment strategies we evaluated, the presented strategy worked best.

The network's input voxel grid is applied to the segment so that its center corresponds to the centroid of the aligned segment.
By default the voxels have minimum side lengths of \unit{0.1}{\meter}.
These can individually be increased to exactly fit segments having one or more larger dimension than the grid.
Whereas maintaining the aspect ratio while scaling can potentially offer better retrieval performance, this individual scaling with a minimum side length better avoids large errors caused by aliasing.
We also found that this scaling method offers the best reconstruction performance, with only a minimal impact on the retrieval performance when the original scale of the segments is passed as a parameter to the network.

\subsection{Training the \segmap{} descriptor}
\label{ssec:training}
In order to achieve both a high retrieval performance and reconstruction capabilities, we propose a customized learning technique.
The two desired objectives are imposed on the network by the \textit{softmax cross entropy loss} $L_c$ for retrieval and the reconstruction loss $L_r$.
We propose to simultaneously apply both losses to the descriptor and to this end define a combined loss function $L$ which merges the contributions of both objectives:
\begin{equation}
L = L_c + \alpha L_r
\end{equation}
where the parameter $\alpha$ weighs the relative importance of the two losses.
The value $\alpha=200$ was empirically found to not significantly impact the performance of the combined network, as opposed to training separately with either of the losses.
Weights are initialized based on Xavier's initialization method (\cite{glorot2010understanding}) and trained using the \ac{ADAM} optimizer (\cite{adam}) with a learning rate of $10^{-4}$.
In comparison to \ac{SGD}, \ac{ADAM} maintains separate learning rates for each network parameter, which facilitates training the network with two separate objectives simultaneously.
Regularization is achieved using dropout (\cite{srivastava2014dropout}) and batch normalization (\cite{ioffe2015batch}).

\textbf{Classification loss} $\boldsymbol{L_c}$ \hspace{3pt}For training the descriptor to achieve better retrieval performance, we use a learning technique similar to the \textit{N-ways classification problem} proposed by~\citet{parkhi2015deep}. 
Specifically, we organize the training data into $N$ classes where each class contains all observations of a segment or of multiple segments that belong to the same object or environment part.
Note that these classes are solely used for training the descriptor and are not related to the semantics presented in Section~\ref{ssec:training_semantics}.
As seen in Fig~\ref{fig:descriptor_architecture}, we then append a classification layer to the descriptor and teach the network to associate a score to each of the $N$ predictors for each segment sample.
These scores are compared to the true class labels using \textit{softmax cross entropy loss}:
\begin{equation}
L_c = -\sum^{N}_{i=1} y_i\log{\frac{e^{l_i}}{\sum^{N}_{k=1} e^{l_k}}}
\end{equation}
where $y$ is the one hot encoded vector of the true class labels and $l$ is the layer output.

Given a large number of classes and a small descriptor dimensionality, the network is forced to learn descriptors that better generalize and prevent overfitting to specific segment samples.
Note that when deploying the system in a new environment the classification layer is removed, as its output is no longer relevant.
The activations of the previous fully connected layer are then used as a descriptor for segment retrieval through \ac{k-NN}.

\textbf{Reconstruction loss} $\boldsymbol{L_r}$ \hspace{3pt}As depicted in Figure~\ref{fig:descriptor_architecture}, map reconstruction is achieved by appending a decoder network and training it simultaneously with the descriptor extractor and classification layer.
This decoder is composed of one fully connected and three deconvolutional layers with a final sigmoid output.
Note that no weights are shared between the descriptor and the decoder networks.
Furthermore, only the descriptor extraction needs to be run in real-time on the robotic platforms, whereas the decoding part can be executed any time a reconstruction is desired.

As proposed by~\citet{brock2016generative}, we use a specialized form of the \textit{binary cross entropy loss}, which we denote by $L_r$:
\begin{equation}
\begin{aligned}
L_r = -\sum_{x,y,z}(&\gamma\, t_{xyz} \log(o_{xyz}) \\
                     &+ (1 - \gamma) (1 - t_{xyz}) \log(1 - o_{xyz}))
\end{aligned}
\end{equation}
where $t$ and $o$ respectively represent the target segment and the network's output and $\gamma$ is a hyperparameter which weighs the relative importance of false positives and false negatives.
This parameter addresses the fact that only a minority of voxels are activated in the voxel grid.
In our experiments, the voxel grids used for training were on average only 3\% occupied and we found $\gamma = 0.9$ to yield good results. 

\subsection{Knowledge transfer for semantic extraction}
\label{ssec:training_semantics}

\begin{figure}
\centering
\includegraphics[width = 0.75\columnwidth]{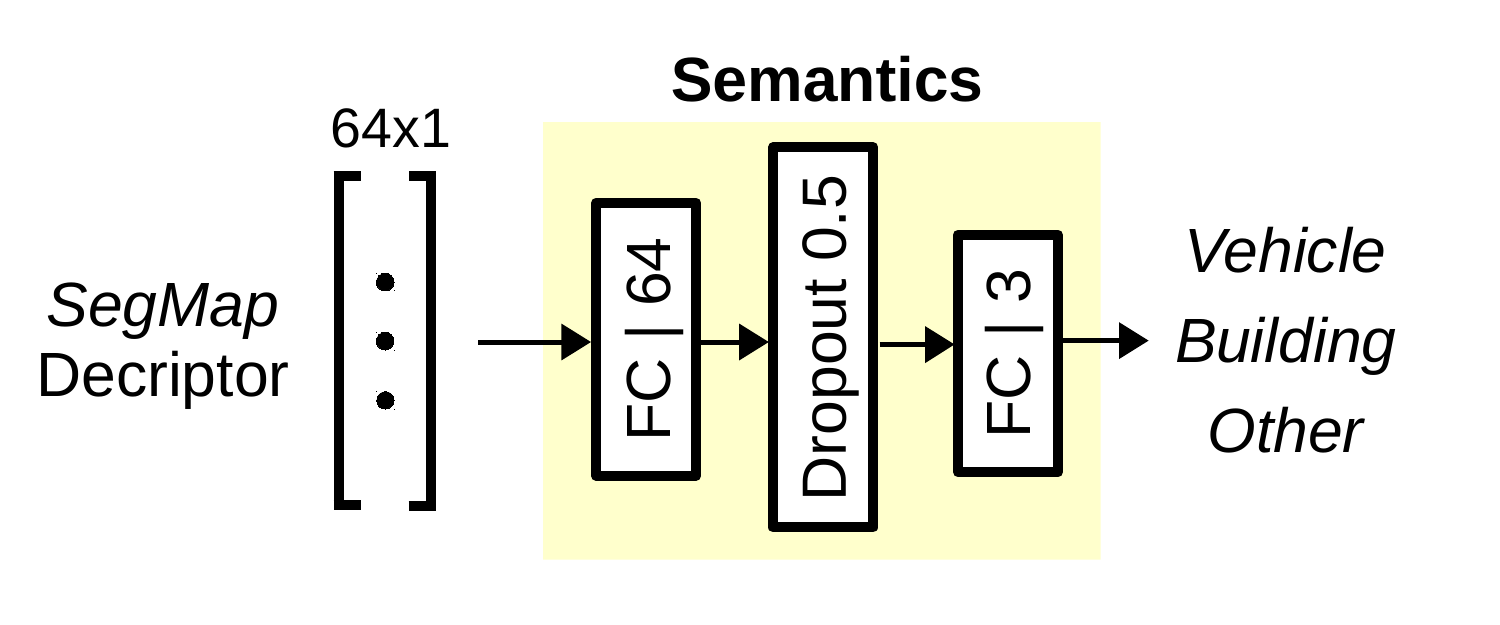}
\caption{A simple fully connected network that can be appended to the \segmap{} descriptor (depicted in Figure~\ref{fig:descriptor_architecture}) in order to extract semantic information.
In our experiments, we train this network to distinguish between vehicles, buildings, and other objects.
}
\label{fig_semantics}
\end{figure}

As can be observed from Figure~\ref{fig_teaser}, segments extracted by the \segmap{} approach for localization and map reconstruction often represent objects or parts of objects.
It is therefore possible to assign semantic labels to these segments and use this information to improve the performance of the localization process.
As depicted in Figure~\ref{fig_semantics}, we transfer the knowledge embedded in our compact descriptor by training a semantic extraction network on top of it.
This last network is trained with labeled data using the \textit{softmax cross entropy loss} and by freezing the weights of the descriptor network.

In this work, we choose to train this network to distinguish between three different semantic classes: \textit{vehicles}, \textit{buildings}, and \textit{others}.
Section~\ref{final_experiment} shows that this information can be used to increase the robustness of the localization algorithm to changes in the environment and to yield smaller map sizes.
This is achieved by rejecting segments associated with potentially dynamic objects, such as vehicles, from the list of segment candidates.

\subsection{\textit{SegMini}}
\label{ssec:segmini}
Finally we propose a lightweight version of the \segmap{} descriptor which is specifically tailored for resource-limited platforms.
\textit{SegMini} has the same architecture as \segmap{} (see Figure~\ref{fig:descriptor_architecture}), with the exception that the number of filter in the convolutional layers and the size of the dense layers is halved. 
Without compromising much on the descriptor retrieval performance this model leads to a computational speedup of 2x for GPU and 6x for CPU (Section~\ref{training_models}).


\begin{figure*}
  \centering
  \includegraphics[width=1.0\linewidth]{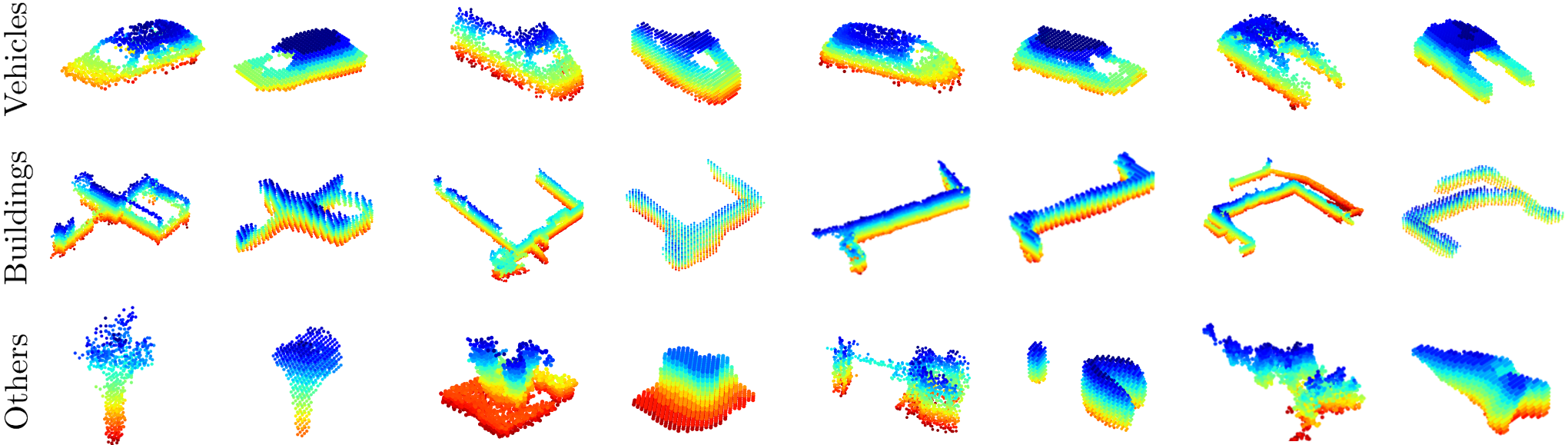}
  \caption{An illustration of the \segmap{} reconstruction capabilities.
  The segments are extracted from sequence 00 of the KITTI dataset and represent, from top to bottom respectively, vehicles, buildings, and other objects. 
  For each segment pair, the reconstruction is shown to the right of the original.
  The network manages to accurately reconstruct the segments despite the high compression to only $64$ values.
  Note that the voxelization effect is more visible on buildings as larger segments necessitate larger voxels to keep the input dimension fixed.}
  \label{fig:reconstructions}
\end{figure*}

\section{EXPERIMENTS}
\label{sec:experiments}

This section presents the experimental validation of our approach.
We first present a procedure for generating training data and detail the performance of the \segmap{} descriptor for localization, reconstruction and semantics extraction.
We finally evaluate the complete \segmap{} solution in multiple real-world experiments.

\subsection{Experiment setup and implementation}
All experiments were performed on a system equipped with an Intel i7-6700K processor, and an NVIDIA GeForce GTX 980 Ti GPU.
The \ac{CNN} models were developed and executed in real-time using the TensorFlow library.
The $libnabo$ library is used for descriptor retrieval with fast \ac{k-NN} search in low dimensional space (\cite{elsebergcomparison}).
The incremental optimization back-end is based on the iSAM2 implementation from \cite{Kaess2012a}.

\subsection{Training data}
\label{ssec:experiments_learning}
The \segmap{} descriptor is trained using real-world data from the KITTI odometry dataset (\cite{geiger2012we}).
Sequences 05 and 06 are used for generating training and testing data, whereas sequence 00 is solely used for validation of the descriptor performance.
In addition, end-to-end experiments are done using sequences 00 and 08, as they feature long tracks with multiple overlapping areas in the trajectories.
For each sequence, segments are extracted using an incremental Euclidean distance-based region growing technique (\cite{dube2018incremental}).
This algorithm extracts point clouds representing parts of objects or buildings which are separated after removing the ground plane (see Figure~\ref{fig:reconstructions}).
The training data is filtered by removing segments with too few observations, or training classes (as described in Section~\ref{ssec:training}) with too few samples. 
In this manner, 3300, 1750, 810 and 2400 segments are respectively generated from sequences 00, 05, 06 and 08, with an average of 12 observations per segment over the whole dataset.

\subsubsection{Data augmentation}
To further increase robustness by reducing sensitivity to rotation and view-point changes in the descriptor extraction process, the dataset is augmented through various transformations at the beginning of each training epoch.
Each segment is rotated at different angles to the alignment described in Section~\ref{sec:alignment} to simulate different view-points.
In order to simulate the effect of occlusion for each segment we remove all points which fall on one side of a randomly generated slicing plane that does not remove more than $50$\% of the points.
Finally, random noise is simulated by randomly removing up to $10$\% of the points in the segment.
Note that these two data augmentation steps are performed prior to voxelization.

\subsubsection{Ground-truth generation}
\label{ssec:ground_truth}
In the following step, we use GPS readings in order to identify ground truth correspondences between segments extracted in areas where the vehicle performed multiple visits.
Only segment pairs with a maximum distance between their centroids of \unit{3.0}{\meter} are considered.
We compute the 3D convex hull of each segment observation $s_1$ and $s_2$ and create a correspondence when the following condition, inspired from the Jaccard index, holds:
\begin{equation}
\frac{\text{Volume}(\text{Conv}(s_1) \cap \text{Conv}(s_2))}{\text{Volume}(\text{Conv}(s_1) \cup \text{Conv}(s_2))} {\geq} p
\end{equation}
In our experiments we found $p = 0.3$ to generate a sufficient number of correspondences while preventing false labelling.
The procedure is performed on sequences 00, 05, and 06, generating 150, 260, and 320 ground truth correspondences respectively.
We use two-thirds of the correspondences for augmenting the training data and one-third for creating validation samples.
Finally, the ground-truth correspondences extracted from sequence 00 are used in Section~\ref{ssec:retrieval_performance} for evaluating the retrieval performance.

\subsection{Training the models}
\label{training_models}
\begin{figure}
  \centering
  \includegraphics[width=1.0\columnwidth]{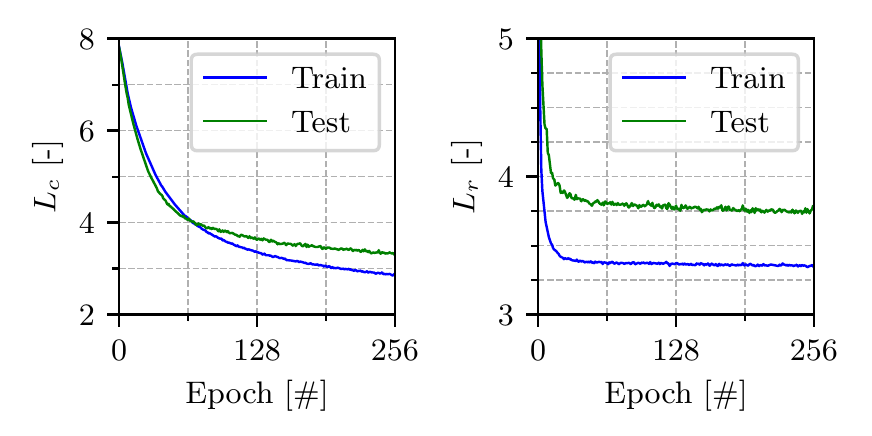}
  \caption{The classification loss $L_c$ (left) and the reconstruction loss $L_r$ (right) components of the total loss $L$, when training the descriptor extractor along with the reconstruction and classification networks.
The depicted reconstruction loss has already been scaled by $\alpha$.}
  \label{fig:loss}
\end{figure}

The descriptor extractor and the decoding part of the reconstruction network are trained using all segments extracted from drive 05 and 06.
Training lasts three to four hours on the GPU and produces the classification and scaled reconstruction losses depicted in Figure~\ref{fig:loss}.
The total loss of the model is the sum of the two losses as describe in Section~\ref{ssec:training}.
We note that for classification the validation loss follows the training loss before converging towards a corresponding accuracy of 41\% and 43\% respectively.
In other words, 41\% of the validation samples were correctly assigned to one of the $N=2500$ classes.
This accuracy is expected given the large quantity of classes and the challenging task of discerning between multiple training samples with similar semantic meaning, but few distinctive features, \eg flat walls.
Note that we achieve a very similar classification loss $L_c$, when training with and without the $L_r$ component of the combines loss $L$.
On a GPU the \segmap{} descriptor takes on average \unit{0.8}{\milli\second} to compute, while the \textit{SegMini} descriptor takes \unit{0.3}{\milli\second}.
On the CPU the performance gain is more significant, as it takes \unit{245}{\milli\second} for a \segmap{} descriptor as opposed to only \unit{41}{\milli\second} for \textit{SegMini}, which is a 6x improvement in efficiency.

\subsection{Descriptor retrieval performance}
\label{ssec:retrieval_performance}
We evaluate the retrieval performance of the \segmap{} descriptor against state-of-the-art methods as well as other networks trained with different secondary goals.
First, our descriptor is compared with eigenvalue-based point cloud features (\cite{weinmann2014semantic}).
We also evaluate the effect of training only for the classification task (Classification) or of training only for the reconstruction one (Autoencoder).
Additionally, we compare classification-based learning with a triplet loss solution (\cite{schroff2015facenet}), where during training, we enforce segments from the same sequence to have a minimal Euclidean distance.
We use a per batch hard mining strategy and the best performing variant of triplet loss as proposed by \cite{hermans2017defense}.
We finally evaluate the \textit{SegMini} model introduced in Section~\ref{ssec:segmini}.

\begin{figure}
  \centering
  \includegraphics{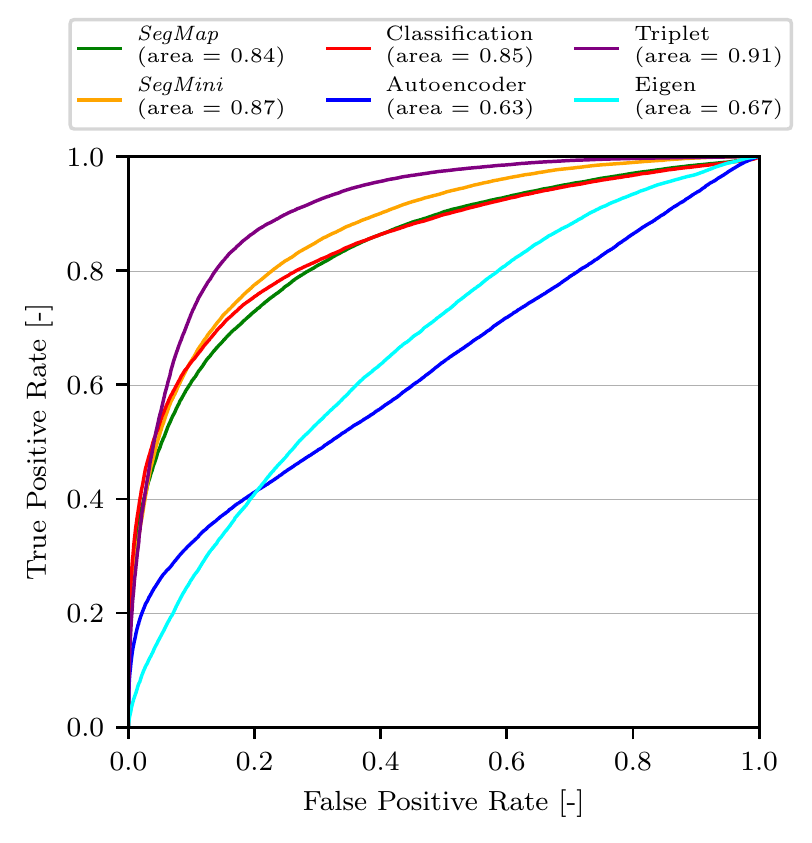}
  \caption{\ac{ROC} curves for the descriptors considered in this work.
  This evaluation is performed using ground-truth correspondences extracted from sequence 00 of the KITTI odometry dataset (\cite{geiger2012we}).
  Note that the \ac{ROC} is not an optimal measure of the quality of the retrieval performance, since it only considers a single threshold for all segment pairs and does not look at the relative ordering of matches on a per query basis.
  }
  \label{fig:roc}
\end{figure}

The retrieval performance of the aforementioned descriptors is depicted in Fig~\ref{fig:roc}.
The \ac{ROC} curves are obtained by generating $45$M labeled pairs of segment descriptors from sequence 00 of the KITTI odometry dataset (\cite{geiger2012we}).
Using ground-truth correspondences, a positive sample is created for each possible segment observation pair.
For each positive sample a thousand negative samples are generated by randomly sampling segment pairs whose centroids are further than \unit{20}{\meter} apart.
The positive to negative sample ratio is representative of our localization problem given that a map created from KITTI sequence 00 contains around a thousand segments.
The \ac{ROC} curves are finally obtained by varying the threshold applied on the $L^2$ norm between the two segment descriptors.  
We note that training with triplet loss offers the best \ac{ROC} performance on these datasets, as it imposes the most consistent separation margin across all segments.

\begin{figure}[t]
  \begin{subfigure}[t]{1.0\columnwidth}
  \centering
  \includegraphics{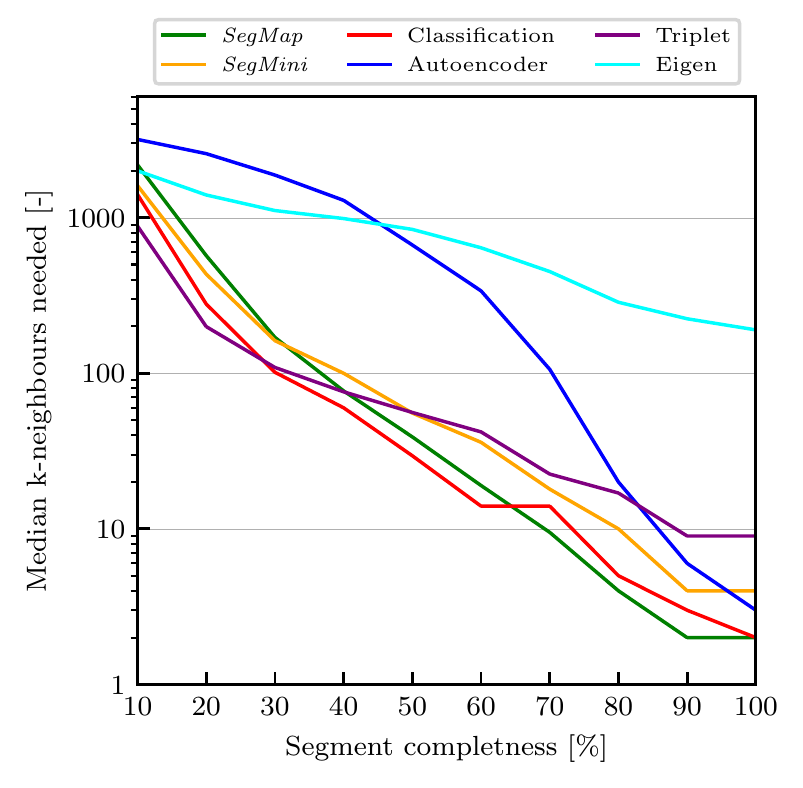}\vspace{-7pt}
  \caption{
  Median k-nearest neighbours needed for all methods as a function of segment completeness.
  }\label{subfig:neighbours-lines}
  \end{subfigure}\hfill\vspace{10pt}
  
  \begin{subfigure}[t]{1.0\columnwidth}
  \centering
  \includegraphics{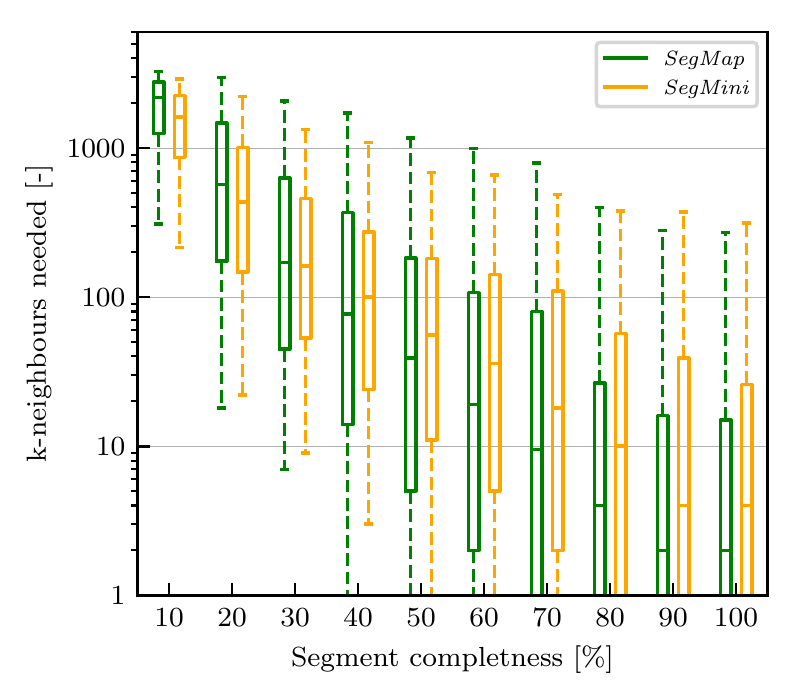}\vspace{-7pt}
  \caption{
  More detailed plot of the k-nearest neighbours needed for the proposed methods as a function of segment completeness.
  }\label{subfig:neighbours-box}
  \end{subfigure}

\caption{
This figure presents how \textit{quickly} descriptors extracted from incrementally grown segments contain relevant information that can be used for localization. 
The x-axis represents the completeness of a segment until all its measurements have been accumulated (here termed \textit{complete}, see Section~\ref{sec:segmap}).
In (\subref{subfig:neighbours-lines}) the log-scaled y-axis represents the median of how many neighbours in the target map need to be considered in order to retrieve the correct target segment (the lower the better).
Similarly (\subref{subfig:neighbours-box}) presents the same results in more detail for the proposed models.
The \segmap{} descriptor offers over the majority of the growing process one order of magnitude better retrieval performance than the hand-crafted baseline descriptor.
}
\label{fig:acc_relative_to_size}
\end{figure}

The \ac{ROC} is not the best evaluation metric for this retrieval task, because it evaluates the quality of classification for a single threshold across all segments.
As introduced in Section~\ref{sec:segmap}, correspondences are made between segments from the local and global maps by using k-NN retrieval in feature space.
The varying parameter is the number of neighbours that is retrieved and not a threshold on the feature distances, which only matter in a relative fashion on a per query basis.
In order to avoid false localizations, the aim is to reduce the number $k$ of neighbours that need to be considered.
Therefore, as a segment grows with time, it is critical that its descriptor converges as quickly as possible towards the descriptor of the corresponding segment in the target map, which in our case is extracted from the last and most \textit{complete} observation (see Section~\ref{sec:segmap}). 
This behaviour is evaluated in Figure~\ref{subfig:neighbours-lines} which relates the number of neighbours which need to be considered to find the correct association, as a function of segment completeness.  
We note that the \segmap{} descriptor offers competitive retrieval performance at every stage of the growing process.
In practice this is important since it allows closing challenging loops such as the one presented in Figure~\ref{fig_teaser}.

Interestingly, the autoencoder has the worst performance at the early growing stages whereas good performance is observed at later stages.
This is in accordance with the capacity of autoencoders to precisely describe the geometry of a segment, without explicitly aiming at gaining a robust representation in the presence of occlusions or changes in view-point.
Although the triplet loss training method offers the best ROC performance, Figure~\ref{subfig:neighbours-lines} suggests that training with the secondry goal of classification yields considerably better results at the later stages of growing.
The poor performance of the triplet loss method especially for very similar segments could be caused by the hard mining amplifying the noise in the dataset.
After a certain point the ordering of matches becomes irrelevant, because the goal is to minimize the number of retrieved neighbours and retrieving too many is computationally unfeasible for later stages of the process.
Therefore although the purely classification-based model performs slightly better for very early observations of a segment, this gain in performance does not matter. 
The proposed \segmap{} descriptor achieves the best performance for very complete segments, where matches are most likely to happen, and maintains a comparable performance across very partial observations.
A more detailed plot for the retrieval performance of the \segmap{} and \textit{SegMini} is presented in Figure~\ref{subfig:neighbours-box}, where also the variance in the retrieval accuracy is shown.

\subsection{Reconstruction performance}

\begin{table}
\fontsize{8}{10}\selectfont
  \renewcommand{\arraystretch}{1.2}
  \centering
  \begin{tabular}{c|c|c|c|c|} \cline{2-5}
    \quad & \multicolumn{4}{|c|}{\textbf{Descriptor size}} \\ 
    \quad & \multicolumn{1}{|c}{16} & \multicolumn{1}{c}{32} & \multicolumn{1}{c}{64} & \multicolumn{1}{c|}{128} \\ \hline
    \multicolumn{1}{|c|}{Autoencoder}        & 0.87 & 0.91 & 0.93 & 0.94 \\ \hline
    \multicolumn{1}{|c|}{\segmap{}} & 0.86 & 0.89 & 0.91 & 0.92 \\ \hline
  \end{tabular}
  \caption{Average ratio of corresponding points within one voxel distance between original and reconstructed segments. 
  Statistics for \segmap{} and the autoencoder baseline using different descriptor sizes.}
  \label{tbl:stats_ae}
\end{table}

In addition to offering high retrieval performance, the \segmap{} descriptor allows us to reconstruct 3D maps using the decoding \ac{CNN} described in Section~\ref{ssec:training}.
Some examples of the resulting reconstructions are illustrated in Figure~\ref{fig:reconstructions}, for various objects captured during sequence 00 of the KITTI odometry dataset.
Experiments done at a larger scale are presented in  Figure~\ref{fig:reconstruction_sar}, where buildings of a powerplant and a foundry are reconstructed by fusing data from multiple sensors.

Since most segments only sparsely model real-world surfaces, they occupy on average only 3\% of the voxel grid.
To obtain a visually relevant comparison metric, we calculate for both the original segment and its reconstruction the ratio of points having a corresponding point in the other segment, within a distance of one voxel.
The tolerance of one voxel means that the shape of the original segment must be preserved while not focusing on reconstructing each individual point.
Results calculated for different descriptor sizes are presented in Table~\ref{tbl:stats_ae}, in comparison with the purely reconstruction focused baseline.
The \segmap{} descriptor with a size of 64 has on average 91\% correspondences between the points in the original and reconstructed segments, and is only slightly outperformed by the autoencoder baseline.
Contrastingly, the significantly higher retrieval performance of the \segmap{} descriptor makes it a clear all-rounder choice for achieving both localization and map reconstruction.

\begin{figure}
  \centering
  \includegraphics[width=1.0\linewidth]{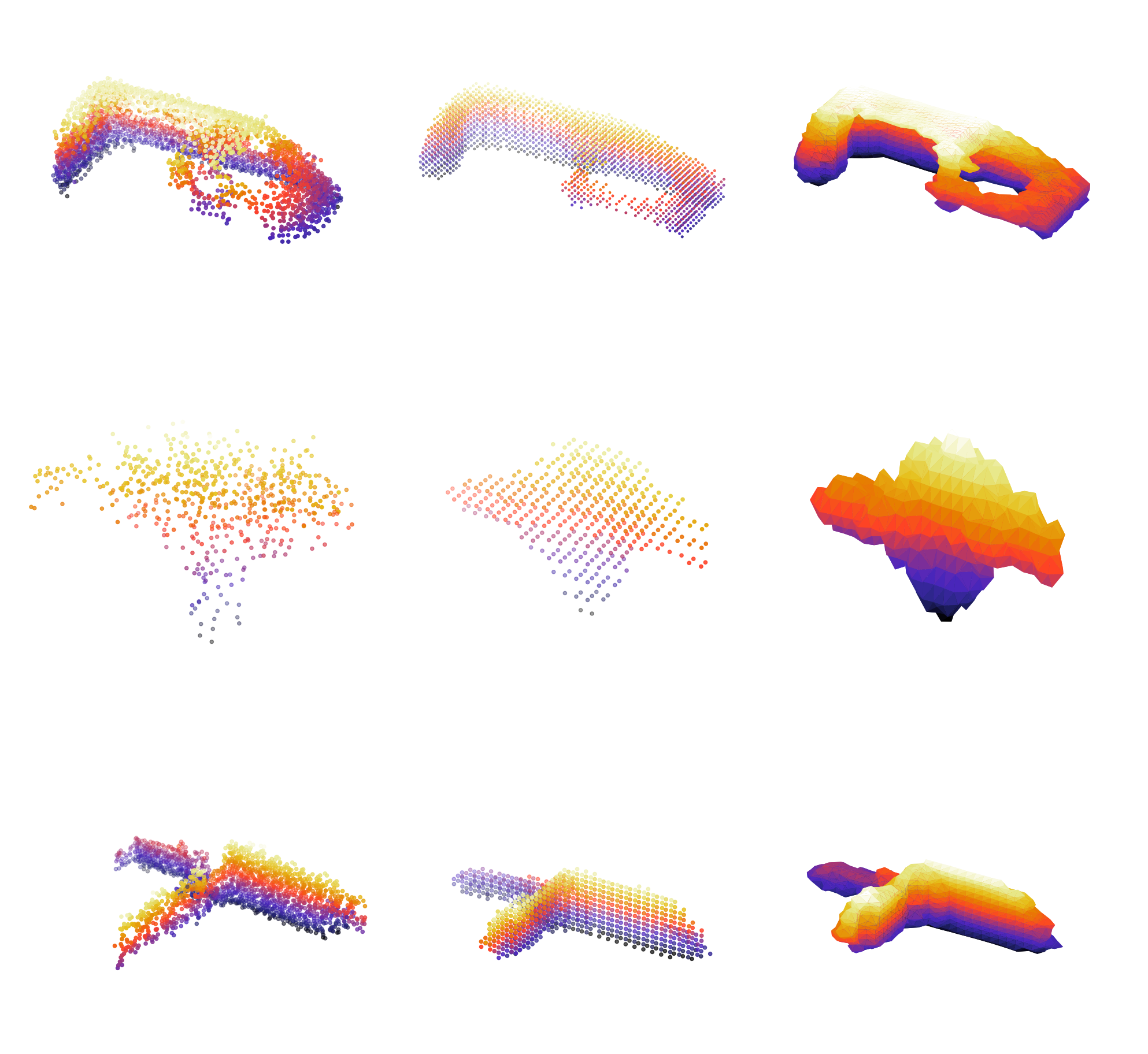}
  \caption{A visual comparison between (left) the original point cloud, (middle) the reconstruction point cloud, and (right) the reconstruction mesh, for 3 segments.}
  \label{fig:reconstruction_meshes}
\end{figure}

\begin{figure*}
  \centering
  \includegraphics[width=0.9\linewidth]{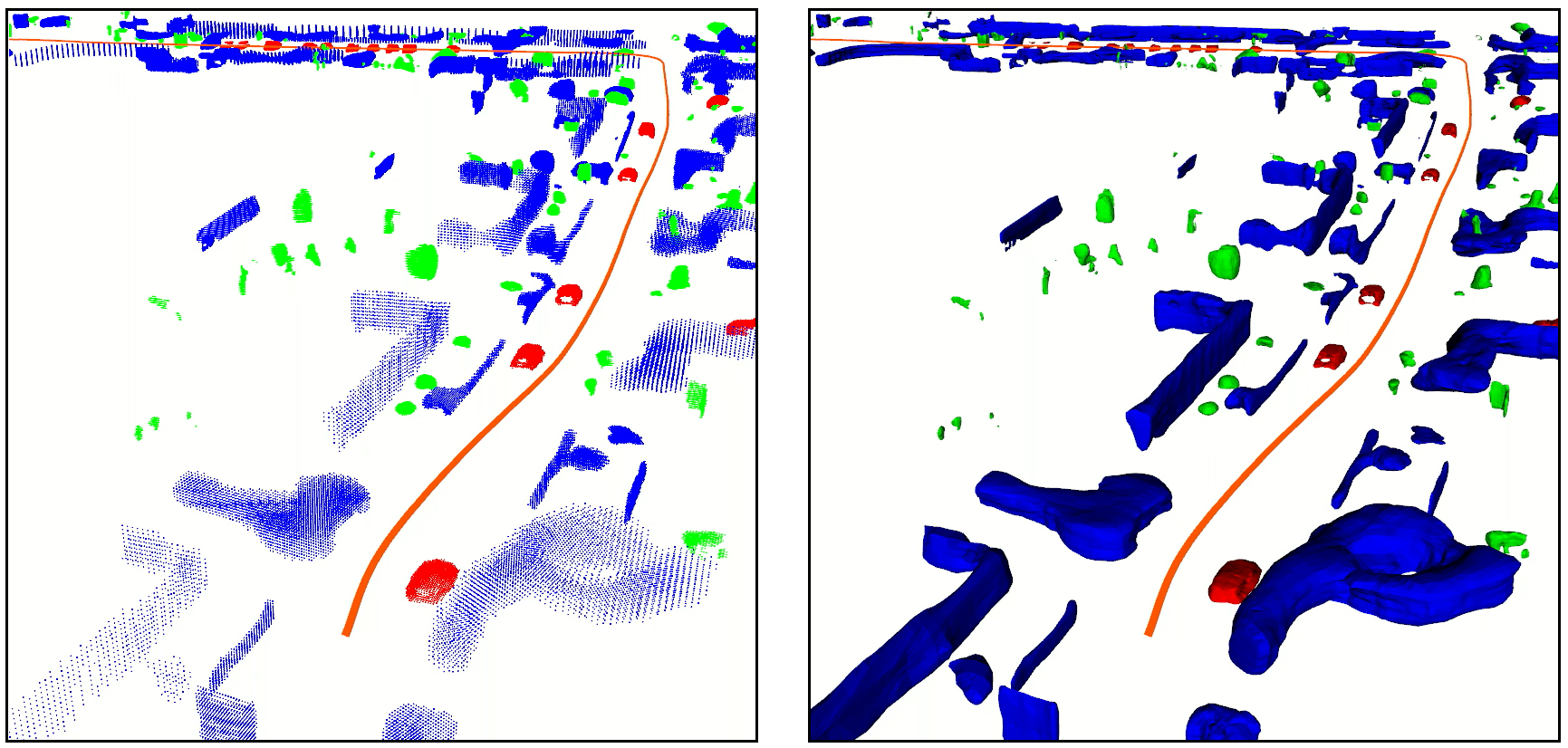}
  \caption{Visualization of segment reconstructions, as point clouds (left), and as surface meshes (right), generated from sequence 00 of the KITTI dataset. The quantization of point cloud reconstructions is most notable in the large wall segments (blue) visible in the background. Equivalent surface mesh representations do not suffer from this issue.}
  \label{fig:reconstruction_meshes_kitti}
\end{figure*}

Overall, the reconstructions are well recognizable despite the high compression ratio.
In Figure~\ref{fig:reconstruction_meshes_kitti}, we note that the quantization error resulting from the voxelization step mostly affects larger segments that have been downscaled to fit into the voxel grid.
%
%
%
%
%
To mitigate this problem, one can adopt a natural approach to representing this information in 3D space, which is to calculate the isosurface for a given probability threshold. 
This can be computed using the ``marching cubes'' algorithm, as presented by \cite{lorensen1987marching}. 
The result is a triangle-mesh surface, which can be used for intuitive visualization, as illustrated in Figure~\ref{fig:reconstruction_meshes} and Figure~\ref{fig:reconstruction_meshes_kitti}. 
%
%

\subsection{Semantic extraction performance}
For training the semantic extractor network (Figure~\ref{fig_semantics}), we manually labeled the last observation of all 1750 segments extracted from KITTI sequence 05.
The labels are then propagated to each observation of a segment for a total of 20k labeled segment observations.
We use 70\% of the samples for training the network and 30\% for validation.
Given the low complexity of the semantic extraction network and the small amount of labeled samples, training takes only a few minutes.
We achieve an accuracy of 89\% and 85\% on the training and validation data respectively.
Note that our goal is not to improve over other semantic extraction methods (\cite{livehicle, qi2016pointnet}), but rather to illustrate that our compressed representation can additionally be used for discarding dynamic elements of the environment and for reducing the map size (Section~\ref{ssec_kitti_slam}).

\subsection{6-DoF pose retrieval performance}
\label{locnet_experiment}

In this section, we demonstrate how the advantageous properties of \segmap{}, particularly the descriptor retrieval performance, translate to state-of-the-art global localization results.
We therefore compare our approach to a global localization method, LocNet (\cite{yin2017locnet, yin2018locnet}).
It uses rotation-invariant, data-driven descriptors that yield reliable matching of 3D LiDAR scans.
LocNet retrieves a nearest neighbor database scan and returns its pose, its output is thus limited to the poses already present in the target map.
Therefore, it works reliably in environments with well defined trajectories (e.g. roads), but fails to return a precise location within large traversable areas such as squares or hallways.
In contrast, \segmap{} uses segment correspondences to estimate an accurate 6-\ac{DoF} pose that includes orientation, which cannot be retrieved directly using the rotation-invariant LocNet descriptors.

\begin{figure}
  \centering
  \includegraphics{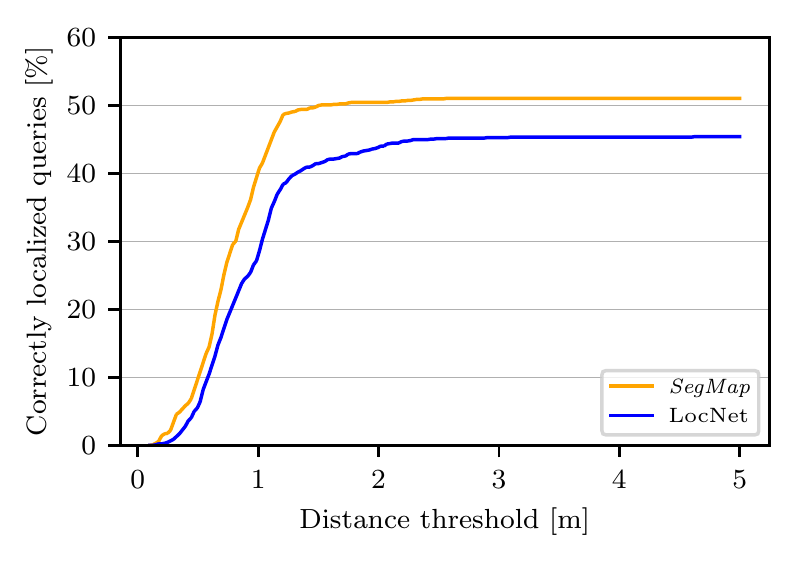}
  \caption{
  Cumulative distribution of position errors on KITTI 00 odometry sequence that compares \segmap{} with state-of-the-art data-driven LocNet approach presented in~\cite{yin2017locnet, yin2018locnet}.
  Our proposed method retrieves a full 6-DoF pose while LocNet uses global scan descriptors to obtain the nearest pose of the target map.
  \segmap{} retrieves poses for a larger number of scans and the returned estimates are more accurate.
  The results saturate at about $52$\% as not all query positions overlap with the target map, with only $65$\% of them being within a radius of \unit{50}{\meter} from the map.
  }
  \label{fig:localizations}
\end{figure}

Figure~\ref{fig:localizations} presents the evaluation of both methods on the KITTI 00 odometry sequence (4541 scans).
We use the first 3000 LiDAR scans and their ground-truth poses to create a map, against which we then localize using the last 1350 scans.
\segmap{} demonstrates a superior performance both by successfully localizing about $6$\% more scans and by returning more accurate localized poses.
To note is that from the query positions only 65\% of them were taken within a distance of \unit{50}{\meter} of the target map, therefore limiting the maximum possible saturation. 
We believe that robust matching of segments, a principle of our method, helps to establish reliable correspondences with the target map, particularly for queries further away from the mapped areas.
This state-of-the-art localization performance is further complemented by a compact map representation, with reconstruction and semantic labeling capabilities.

\subsection{A complete mapping and localization system}
\label{loam_experiment}

\begin{figure*}[t]
  \begin{subfigure}[t]{1.0\columnwidth}
  \centering
  \includegraphics{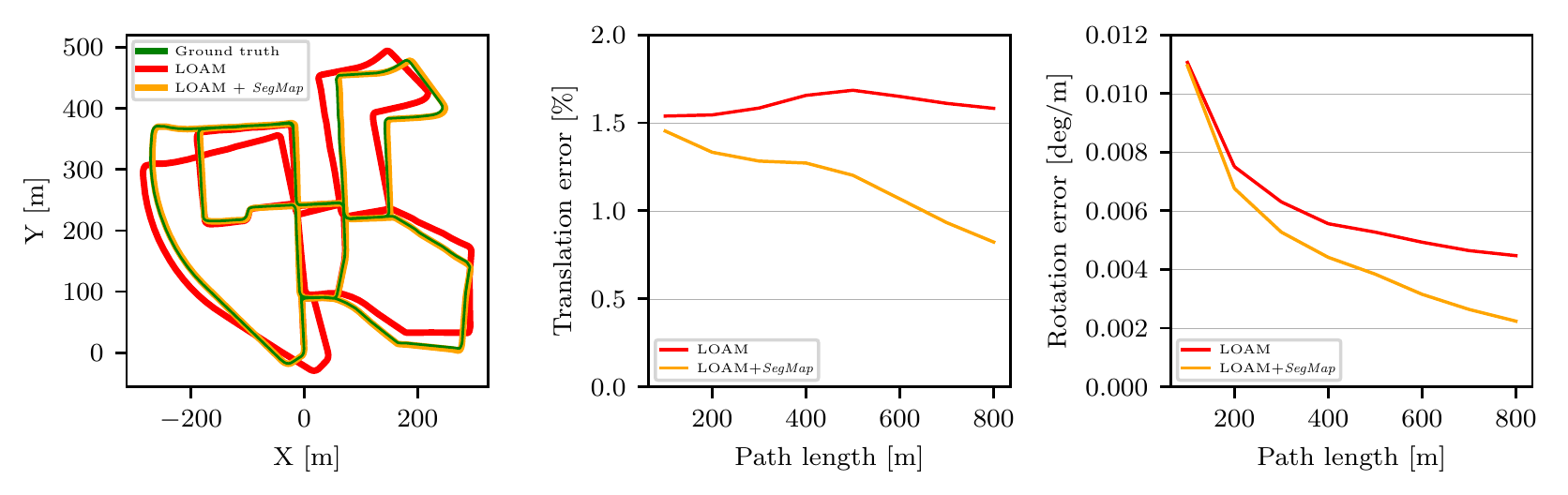}\vspace{-7pt}
  \caption{
  KITTI odometry sequence 00.
  }\label{subfig:trajectory-00}
  \end{subfigure}\hfill\vspace{10pt}
  
  \begin{subfigure}[t]{1.0\columnwidth}
  \centering
  \includegraphics{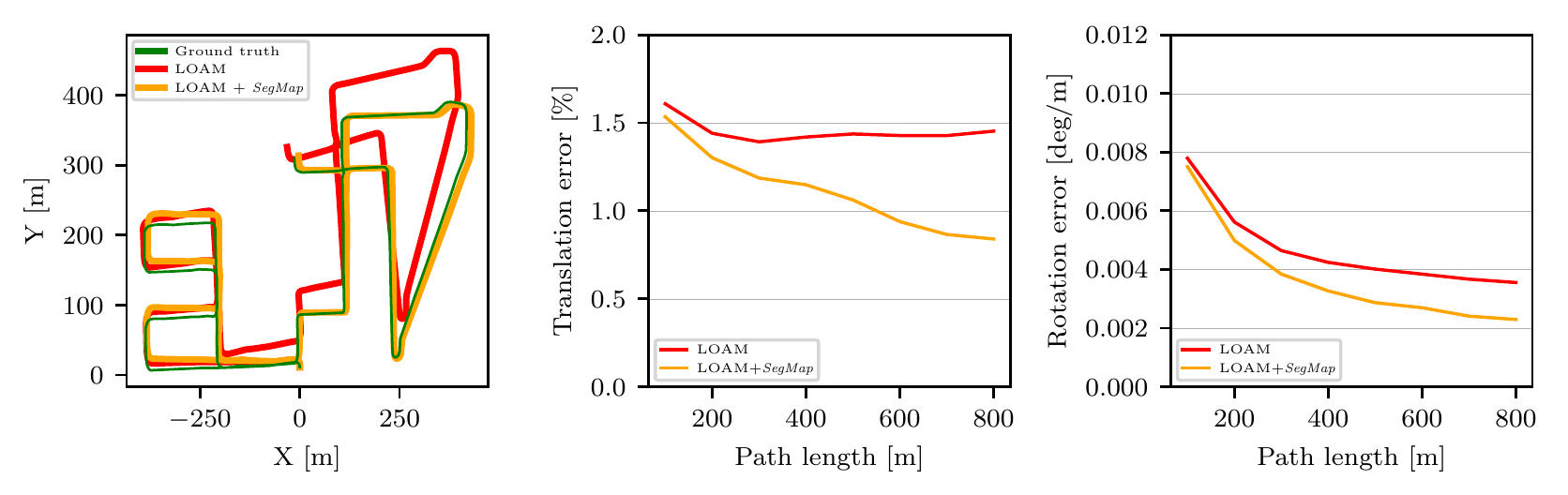}\vspace{-7pt}
  \caption{
  KITTI odometry sequence 08.
  }\label{subfig:trajectory-08}
  \end{subfigure}

\caption{
The trajectories for KITTI odometry sequences \subref{subfig:trajectory-00})~00 and \subref{subfig:trajectory-08})~08 for LOAM and the combination of LOAM and \segmap{}. In addition we show translation and rotation errors for the two approaches, using the standard KITTI evaluation method~\cite{geiger2012we}.
}
  \label{fig:segloam}
\end{figure*}

So far, we have only evaluated \segmap{} as a stand-alone global localization system, demonstrating the performance of segment descriptors and the 6-\ac{DoF} pose retrieval.
Such global localization systems, however, are commonly used in conjunction with odometry and mapping algorithms.
To prove the qualities of \segmap{} in such a scenario, we have combined it with a state-of-the-art LiDAR odometry and mapping system, LOAM (\cite{zhang2014loam}).
Our implementation is based on a publicly available version of LOAM and achieves similar odometry performance results on KITTI, as the ones reported by other works, such as \cite{velas2018cnn}.
We use a loosely coupled approach, where LOAM is used to undistort the scans and provide an odometry estimate between frames, in real-time.
The scans from LOAM are used to build a local map from which segments are extracted and attached to a pose-graph, together with the odometry measurements.
Loop closures can then be added in real-time as constraints in the graph, to correct the drifting odometry.
This results in a real-time LiDAR-only end-to-end pipeline that produces segment-based maps of the environment, with loop-closures.

In all experiments, we use a local map with a radius of \unit{50}{\meter} around the robot.
When performing segment retrieval we consider $64$ neighbours  and require a minimum of $7$ correspondences, which are altogether geometrically consistent, to output a localization.
These parameters were chosen empirically using the information presented in Figure~\ref{fig:roc}~and~\ref{fig:acc_relative_to_size} as a reference.

Our evaluations on KITTI sequences 00 and 08 (Figure~\ref{fig:segloam}) demonstrate that global localization results from \segmap{} help correct for the drift of the odometry estimates.
The trajectories outputted by the system combining \segmap{} and LOAM, follow more precisely the ground-truth poses provided by the benchmark, compared to the open-loop solution.
We also show how global localizations reduce both translational and rotational errors.
Particularly over longer paths \segmap{} is able to reduce the drift in the trajectory estimate by up to 2 times, considering both translation and rotation errors.
For shorter paths, the drift only improves marginally or remains the same, as local errors are more dependent on the quality of the odometry estimate.
We believe that our evaluation showcases not only the performance of \segmap{}, but also the general benefits stemming from global localization algorithms.

\subsection{Multi-robot experiments}

\label{final_experiment}
We evaluate the \segmap{} approach on three large-scale multi-robot experiments: one in an urban-driving environment and two in search and rescue scenarios.
In both indoor and outdoor scenarios we use the same model which was trained on the KITTI sequences 05 and 06 as described in Section~\ref{training_models}.

The experiments are run on a single machine, with a multi-thread approach to simulating a centralized system.
One thread per robot accumulates the 3D measurements, extracting segments, and performing the descriptor extraction.
The descriptors are transmitted to a separate thread which localizes the robots through descriptor retrieval and geometric verification, and runs the pose-graph optimization. 
In all experiments, sufficient global associations need to be made, in real-time, for co-registration of the trajectories and merging of the maps. 
Moreover in a centralized setup it might be crucial to limit the transmitted data over a wireless network with potentially limited bandwidth.

\subsubsection{Multi-robot SLAM in urban scenario}
\label{ssec_kitti_slam}
In order to simulate a multi-robot setup, we split sequence 00 of the KITTI odometry dataset into five sequences, which are simultaneously played back on a single computer for a duration of 114 seconds.
%
In this experiment, the semantic information extracted from the \segmap{} descriptors is used to reject segments classified as \textit{vehicles} from the retrieval process.

With this setup, $113$ global associations were discovered, allowing to link all the robot trajectories and create a common representation.
We note that performing \ac{ICP} between the associated point clouds would refine the localization transformation by, on average, only \unit{0.13\pm0.06}{\meter} which is in the order of our voxelization resolution. 
However, this would require the original point cloud data to be kept in memory and transmitted to the central computer. 
Future work could consider refining the transformations by performing \ac{ICP} on the reconstructions.

Localization and map reconstruction was performed at an average frequency of \unit{10.5}{\hertz} and segment description was responsible for $30\%$ of the total runtime with an average duration of \unit{28.4}{\milli\second} per local cloud.
A section of the target map which has been reconstructed from the descriptors is depicted in Figure~\ref{fig_teaser}.

Table~\ref{tbl:stats_experiment} presents the results of this experiment.
The required bandwidth is estimated by considering that each point is defined by three 32-bit floats and that 288 additional bits are required to link each descriptor to the trajectories.
We only consider the \textit{useful data} and ignore any transfer overhead.
The final map of the KITTI sequence 00 contains $1341$ segments out of which $284$ were classified as vehicles.
A map composed of all the raw segment point clouds would be $16.8$ MB whereas using our descriptor it is reduced to only $386.2$ kB.
This compression ratio of $43.5$x can be increased to $55.2$x if one decides to remove vehicles from the map.
%
%
This shows that our approach can be used for mapping much larger environments. 

\begin{table}
\fontsize{6}{8}\selectfont
  \renewcommand{\arraystretch}{1.2}
  \centering
  \caption{Statistics resulting from the three experiments.}
  \label{tbl:stats_experiment}
  \resizebox{1.0\columnwidth}{!}{%
  \begin{tabular}{|l|ccc|} \hline
    \textbf{Statistic} & \textbf{KITTI} & \textbf{Powerplant} & \textbf{Foundry} \\ \hline
    Duration (s)  & 114 & 850  & 1086 \\ \hline
    Number of robots  & 5 & 3  & 2 \\ \hline
    Number of segmented local cloud  & 557 & 758  & 672 \\ \hline
    Average  number of segments per cloud & 42.9 & 37.0  & 45.4 \\ \hline
    Bandwidth for transmitting local clouds (kB/s) & 4814.7 & 1269.2  & 738.1 \\ \hline
	Bandwidth for transmitting segments (kB/s) & 2626.6 & 219.4  & 172.2 \\ \hline    
    Bandwidth for transmitting descriptors (kB/s) & 60.4 & 9.5  & 8.1 \\ \hline
    Final map size with the \segmap{} descriptor (kB) & 386.2 & 181.3  & 121.2 \\ \hline
    Number of successful localizations & 113 & 27 & 85 \\ \hline
  \end{tabular}}
\end{table}


\subsubsection{Multi-robot SLAM in disaster environments}
\begin{figure}
  \centering
  \includegraphics[width=1.0\columnwidth]{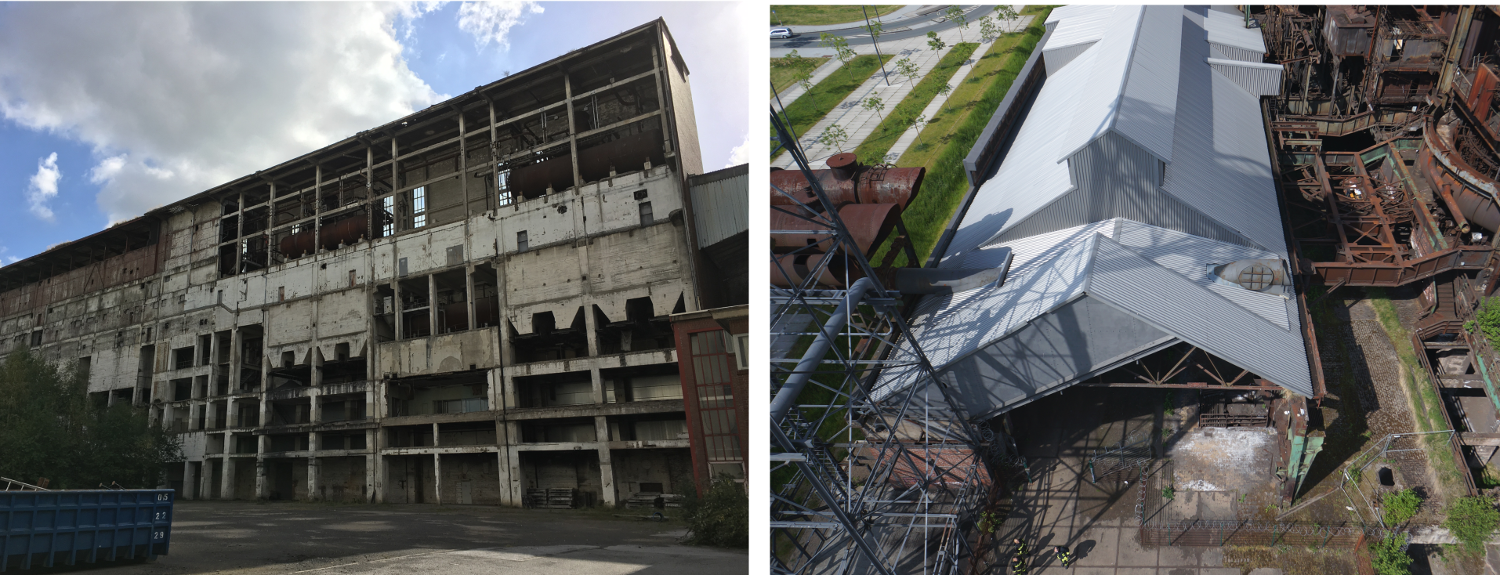}
  \caption{Buildings of the Gustav Knepper powerplant (left) and the Phoenix-West foundry (right).}
  \label{fig:environments}
\end{figure}

For the two following experiments, we use data collected by \acp{UGV} equipped with multiple motor encoders, an Xsens MTI-G \ac{IMU} and a rotating 2D SICK LMS-151 LiDAR.
First, three \acp{UGV} were deployed at the decommissioned Gustav Knepper powerplant: a large two-floors utility building measuring \unit{100}{\meter} long by \unit{25}{\meter} wide.
The second mission took place at the Phoenix-West foundry in a semi-open building made of steel.
A section measuring \unit{100}{\meter} by \unit{40}{\meter} was mapped using two \acp{UGV}.
The buildings are shown in Fig~\ref{fig:environments}.
%

\begin{figure}
  \centering
  \includegraphics[width=1.0\columnwidth]{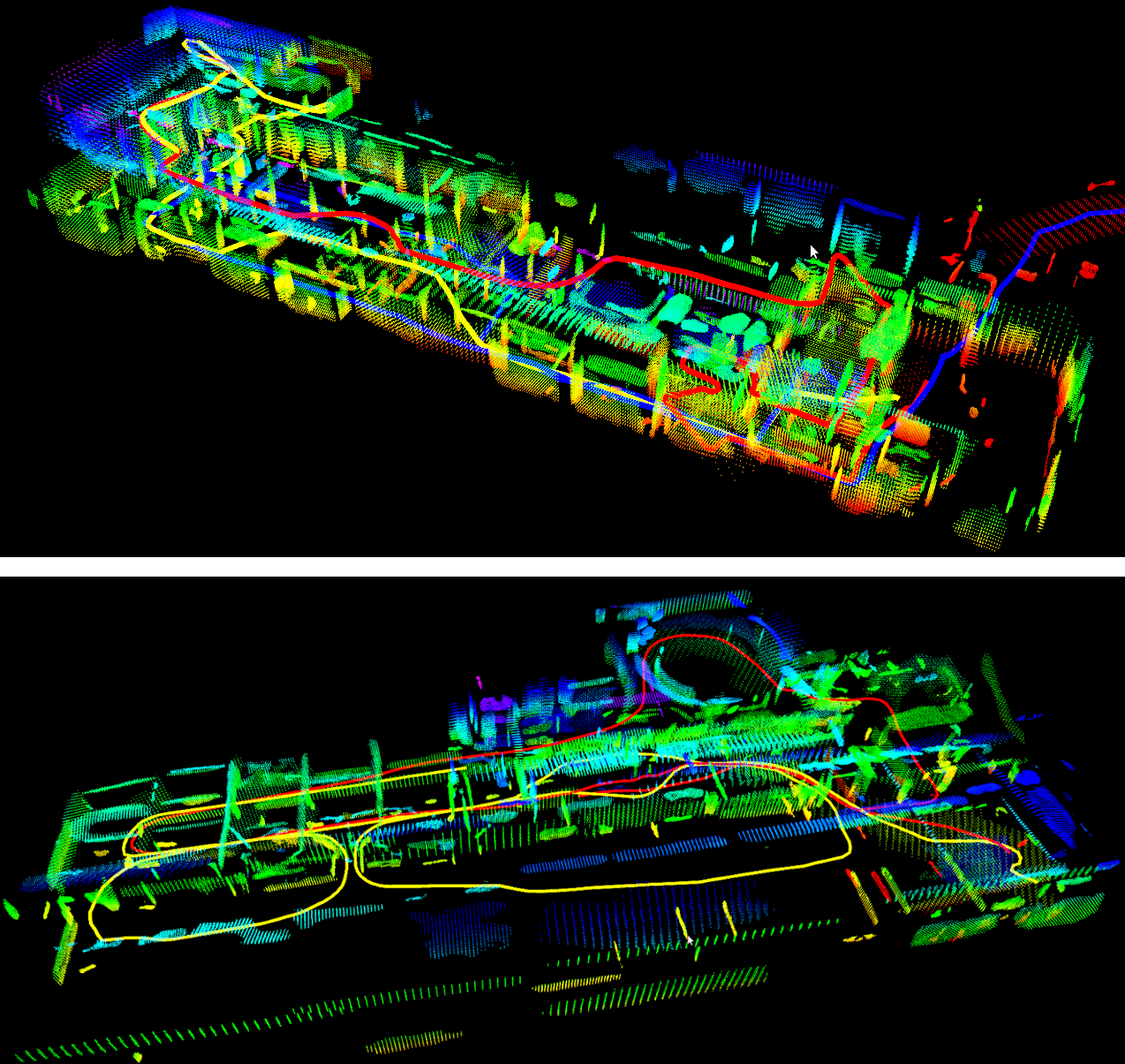}
  \caption{This figure illustrates a reconstruction of the buildings of the Gustav Knepper powerplant (top) and the Phoenix-West foundry (bottom).
  The point clouds are colored by height and the estimated robot trajectories are depicted with colored lines.
  }
  \label{fig:reconstruction_sar}
\end{figure}

For these two experiments, we used an incremental smoothness-based region growing algorithm which extracts plane-like segments (\cite{dube2018incremental}).
The resulting \segmap{} reconstructions are shown in  Figure~\ref{fig:reconstruction_sar} and detailed statistics are  presented in Table~\ref{tbl:stats_experiment}.
Although these planar segments have a very different nature than the ones used for training the descriptor extractor, multiple localizations have been made in real-time so that consistent maps  could be reconstructed in both experiments. 
Note that these search and rescue experiments were performed with sensors without full 360\degree{} field of view.
Nevertheless, \segmap{} allowed robots to localize in areas visited in opposite directions.

\section{DISCUSSION AND FUTURE WORK}
\label{sec:discussion}

While our proposed method works well in the demonstrated experiments it is limited by the ability to only observe the geometry of the surrounding structure.
This can be problematic in some man-made environments, which are repetitive and can lead to perceptual aliasing, influencing both the descriptor and the geometric consistency verification.
This could be addressed by detecting such aliasing instances and dealing with them explicitly, through for example an increase in the constraints of the geometric verification.
On the other hand, featureless environments, such as for example flat fields or straight corridors, are equally challenging.
As even LIDAR-based odometry methods struggle to maintain an accurate estimate of pose, these environments do not allow for reliable segment extraction.
In these cases the map will drift until a more distinct section is reached that can be loop-closed, thus allowing for partial correction of the previously built pose-graph.
In different environments the two segmentation algorithms will have varying performances, with the Euclidean distance based one working better in outdoor scenarios, while the curvature-based one is more suited for indoor scenarios.
A future approach would be to run the two segmentation strategies in parallel, thus allowing them to compensate for each others short-comings and enabling robots to navigate in multiple types of environments during the same mission.

In order to address some of the aforementioned drawbacks, in future work we would like to extend the \segmap{} approach to different sensor modalities and different point cloud segmentation algorithms.
For example, integrating information from camera images, such as color, into the descriptor learning could mitigate the lack of descriptiveness of features extracted from segments with little distinct geometric structure.
In addition, color and semantic information from camera images could not only be used to improve the descriptor but also to enhance the robustness of the underlying segmentation process.
Considering the real-time constraints of the system, to note with respect to future work are the additional computational expenses introduced by processing and combining more data modalities. 

Furthermore, whereas the present work performs segment description in a discrete manner, it would be interesting to investigate incremental updates of learning-based descriptors that could make the description process more efficient, such as the voting scheme proposed by~\citet{engelcke2017vote3deep}.
Instead of using a feed-forward network, one could also consider a structure that leverages temporal information in the form of recurrence in order to better describe segments based on their evolution in time.
Moreover, it could be of interest to learn the usefulness of segments as a precursory step to localization, based on their distinctiveness and semantic attributes.

\section{CONCLUSION}
\label{sec:conclusion}
This paper presented \segmap{}: a segment-based approach for \textit{map representation} in localization and mapping with 3D sensors.
In essence, the robots' surroundings are decomposed into a set of segments, and each segment is represented by a distinctive, low dimensional learning-based descriptor.
Data associations are identified by segment descriptor retrieval and matching, made possible by the repeatable and descriptive nature of segment-based features.

We have shown that the descriptive power of \segmap{} outperforms hand-crafted features as well as the evaluated data-driven baseline solutions.
Our experiments indicate that \segmap{} offers competitive localization performance, in comparison to the state-of-the-art LocNet method.
Additionally, we have combined our localization approach with LOAM, a LiDAR-based local motion estimator, and have demonstrated that the output of \segmap{} helps correct the drift of the open-loop odometry estimate.
Finally, we have introduced \textit{SegMini}: a light-weight version of our \segmap{} descriptor which can more easily be deployed on platforms with limited computational power. 

In addition to enabling global localization, the \segmap{} descriptor allows us to reconstruct a map of the environment and to extract semantic information.
The ability to reconstruct the environment while achieving a high compression rate is one of the main features of \segmap{}.
This allows us to perform both \ac{SLAM} and 3D reconstruction with LiDARs at large scale and with low communication bandwidth between the robots and a central computer.
These capabilities have been demonstrated through multiple experiments with real-world data in urban driving and search and rescue scenarios.
The reconstructed maps could allow performing navigation tasks such as, for instance, multi-robot global path planning or increasing situational awareness.

\section*{ACKNOWLEDGMENTS}
This work was supported by the European Union's Seventh Framework Program for research, technological development and demonstration under the TRADR project No. FP7-ICT-609763, from the EU H2020 research project under grant agreement No 688652, the Swiss State Secretariat for Education, Research and Innovation (SERI) No 15.0284, and by the Swiss National Science Foundation through the National Center of Competence in Research Robotics (NCCR).
The authors would like to thank Abel Gawel, Mark Pfeiffer, Mattia Gollub, Helen Oleynikova, Philipp Kr\"usi, Igor Gilitschenski and Elena Stumm for their valuable collaboration and support. 


\bibliographystyle{plainnat}
\bibliography{ijrr.bib}

\acrodefplural{CNN}[CNNs]{Convolutional Neural Networks}
\acrodefplural{UGV}[UGVs]{Unmanned Ground Vehicles}
\acrodefplural{ReLU}[ReLUs]{Rectified Linear Units}

\begin{acronym}
\acro{ICP}{Iterative Closest Point}
\acro{MAP}{Maximum A Posteriori}
\acro{SLAM}{Simultaneous Localization and Mapping}
\acro{DoF}{Degree-of-Freedom}
\acro{GUI}{Graphical User Interface}
\acro{TRADR}{``Long-Term Human-Robot Teaming for Robots Assisted Disaster Response''}
\acro{SaR}{Search and Rescue}
\acro{UGV}{Unmanned Ground Vehicle}
\acro{IMU}{Inertial Measurement Unit}
\acro{k-NN}{k-Nearest Neighbors}
\acro{FPFH}{Fast Point Feature Histograms}
\acro{CNN}{Convolutional Neural Network}
\acro{ReLU}{Rectified Linear Unit}
\acro{ROC}{Receiver Operating Characteristic}
\acro{ADAM}{Adaptive Moment Estimation}
\acro{SGD}{Stochastic Gradient Descent}
\acro{TPR}{True Positive Rate}
\acro{FPR}{False Positive Rate}
\end{acronym}

\end{document}